\documentclass{article}
\usepackage{ijcai25}

\usepackage{times}
\usepackage{soul}
\usepackage{url}
\usepackage[hidelinks]{hyperref}
\usepackage[utf8]{inputenc}
\usepackage[small]{caption}
\usepackage{graphicx}
\usepackage{amsmath}
\usepackage{amsthm}
\usepackage{booktabs}
\usepackage{algorithm}
\usepackage{algorithmic}
\usepackage[switch]{lineno}

\usepackage{booktabs}
\usepackage{multirow}
\usepackage{amssymb}
\usepackage{makecell}

\newtheorem{definition}{Definition}

\usepackage[table]{xcolor}
\usepackage{xcolor}

\title{Recent Advances in Out-of-Distribution Detection with CLIP-Like Models: A Survey}

\author{
Chaohua Li$^{1,2}$
\and
Enhao Zhang$^{1,2}$\and
Chuanxing Geng$^{1,2,3}$\And
Songcan Chen$^{1,2}$\thanks{Corresponding author.}\\
\affiliations
$^1$College of Computer Science and Technology, Nanjing University of Aeronautics and Astronautics\\
$^2$MIIT Key Laboratory of Pattern Analysis and Machine Intelligence\\
$^3$Department of Computer Science, Hong Kong Baptist University\\
\emails
\{chaohuali, zhangeh, gengchuanxing, s.chen\}@nuaa.edu.cn
}

\begin{document}

\maketitle

\begin{abstract}
    Out-of-distribution detection (OOD) is a pivotal task for real-world applications that trains models to identify samples distributionally different from the in-distribution (ID) data during testing. Recent advances in AI, particularly Vision-Language Models (VLMs) like CLIP, have revolutionized OOD detection by shifting from traditional unimodal image detectors to multimodal image-text detectors. This shift has inspired extensive research, however, existing categorization scheme (\textit{e.g.,} few$/$zero-shot types) still rely solely on the availability of \textit{ID images}, following a unimodal paradigm. To better align with CLIP’s cross-modal nature, we propose a new categorization scheme rooted in both \textit{image} and \textit{text modalities}. Specifically, we categorize existing methods guided by how the visual and textual information of \textit{OOD data} is separately utilized in image $+$ text modalities, further dividing them into four groups: \{OOD Images (\textit{i.e.,} outliers) Seen$/$Unseen $+$ OOD Texts (\textit{i.e.,} learnable vectors or class names) Known$/$Unknown\} across two training strategies (\textit{e.g.,} train-free$/$required). More importantly, we further discuss the open problems of CLIP-like OOD detection, and highlight potential directions for future research, including cross-domain integration, practical applications, and theoretical understanding. 
\end{abstract}

\section{Introduction}

\begin{figure}[t]
\centering
\includegraphics[width=0.927 \linewidth]{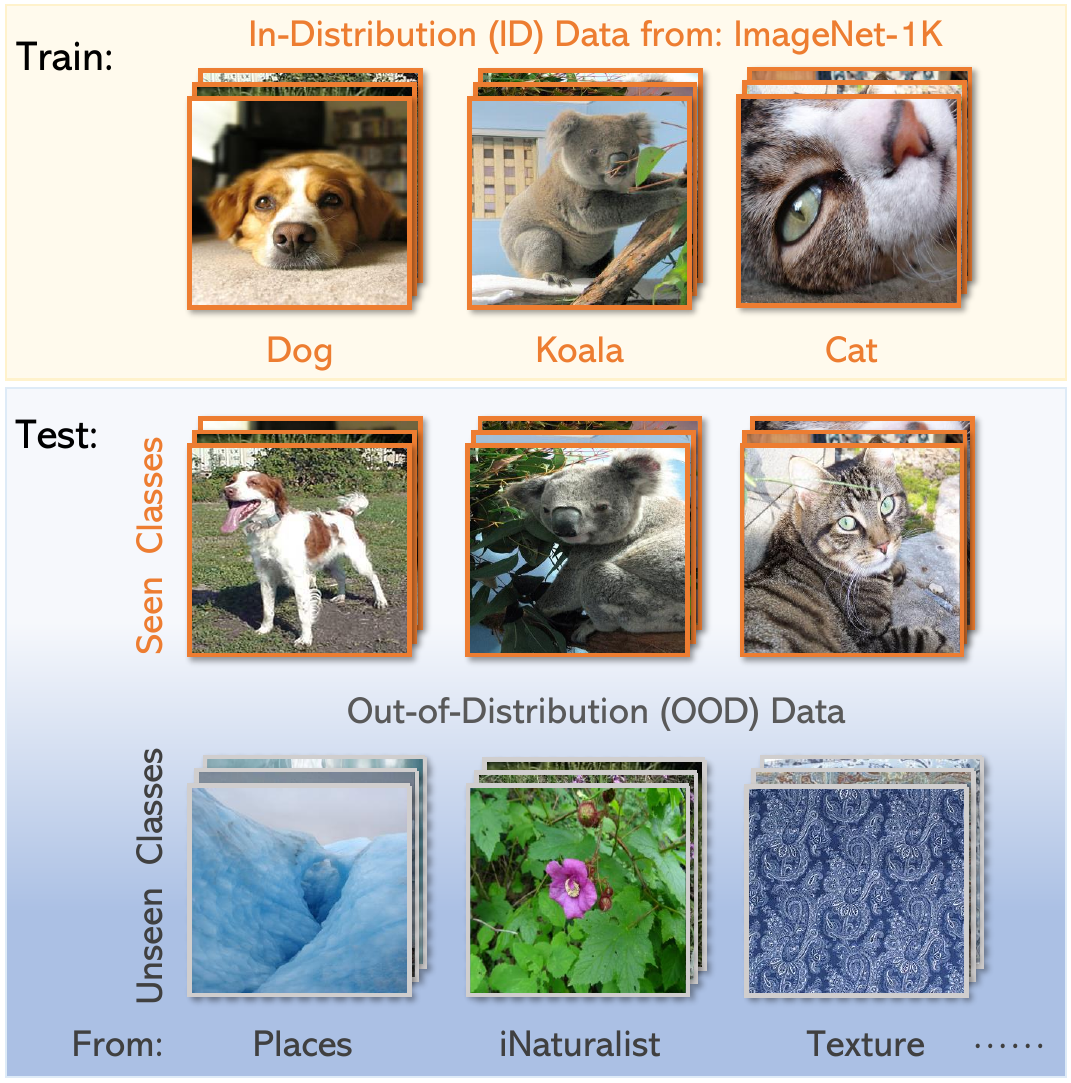} 
\caption{Illustration of Out-of-distribution (OOD) Detection. OOD detection is a major computer vision task that addresses semantic distribution shifts between training and testing data, reflecting real-world scenarios. Specifically, during training, the model learns from ID data, as shown in the upper. However, once deployed, the model may encounter both ID (\textit{\textcolor{orange}{seen}}) and OOD (\textit{\textcolor{gray}{unseen}}) data in the testing phase, as depicted in the below. To ensure reliability, the trained model should not only classify ID samples accurately but also detect OOD instances to avoid incorrect decisions.}
\label{fig1}
\end{figure}

Recent advances in artificial intelligence (AI) have enabled modern systems to perform exceptionally well under laboratory-level environments \cite{russakovsky2015imagenet,he2015delving}. These systems are typically trained following the \textit{closed-set} assumption, where both the training and testing data are drawn from the identical label and feature spaces (\textit{i.e.,} in-distribution (ID)) \cite{scheirer2012toward}. However, this assumption does not always hold in real-world scenarios, as it is infeasible to include all possible data in training without omissions. As a result, a robust AI system should possess not only the ability to classify ID data accurately, but also handle out-of-distribution (OOD) data effectively in real-time, as shown in Fig.~\ref{fig1}, \cite{hendrycks2016baseline,huang2021mos}. For example, an autonomous driving system should be able to alert the driver to unidentified obstacles instead of misclassifying them as ordinary objects, which could lead to catastrophic consequences \cite{huang2020survey}. Similarly, in medical diagnostics \cite{hong2024out}, an AI-powered system must be adept at detecting anomalies such as lesions without confusing them with normal samples, ensuring accurate and reliable diagnoses. Moreover, detecting OOD data is crucial in tasks such as novel species detection in biodiversity studies \cite{marsland2003novelty} and lifelong learning systems \cite{saar2013lifelong} that continuously adapt to new knowledge.

\begin{figure}[t]
\centering
\includegraphics[width=0.9 \linewidth]{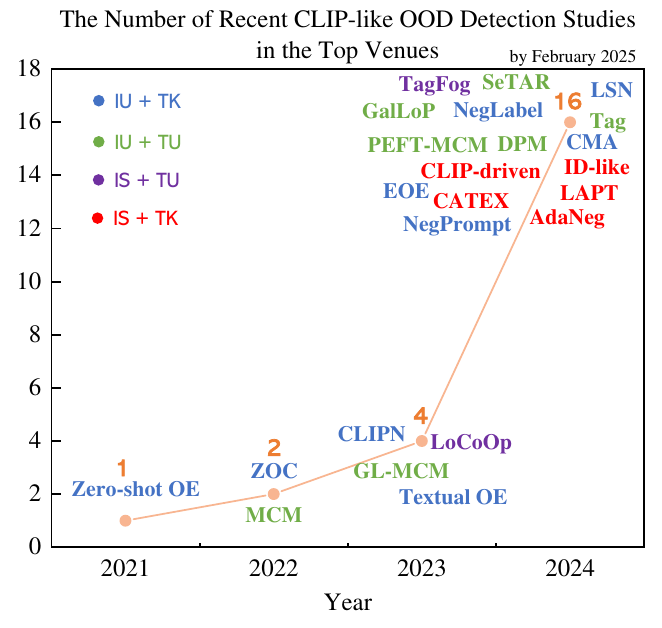} 
\caption{The number of recent CLIP-like OOD detection studies in top venues (up to 02/2025). The abbreviations \textcolor{blue}{IU + TK}, \textcolor{green}{IU + TU}, \textcolor{purple}{IS + TU} and \textcolor{red}{IS + TK}, represent OOD Images Unseen + OOD Texts Known, OOD Images Unseen + OOD Texts Unknown, OOD Images Seen + OOD Texts Unknown and OOD Images Seen + OOD Texts Known, respectively.}
\label{fig2}
\end{figure}

\begin{figure*}[t]
\centering
\includegraphics[scale=0.5]{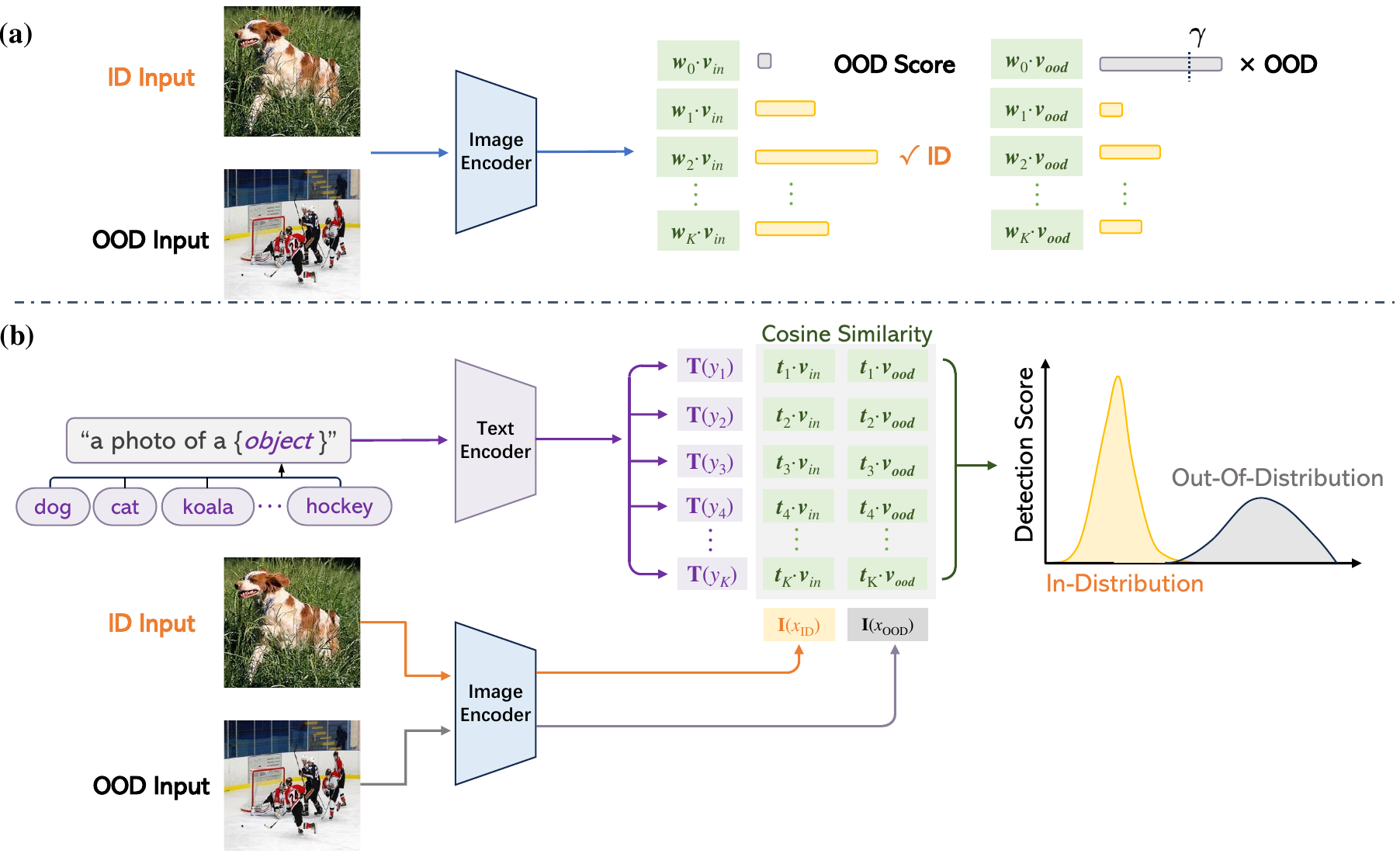} 
\caption{The general pipeline of unimodal and CLIP-like OOD detection paradigms. \textbf{(a)} Unimodal OOD detection models typically use a single-image encoder trained on ID data. The encoder extracts visual embeddings and often employs confidence scores, distance metrics, or density estimation with a threshold. \textbf{(b)} Compared to the unimodal ones, CLIP learns a joint vision-language embedding space by aligning images with textual descriptions. This allows CLIP to map diverse visual concepts into a more semantically structured embedding space.}
\label{fig3}
\end{figure*}

Out-of-distribution (OOD) detection has garnered significant attention in recent years. Traditionally, OOD detection methods were primarily developed within a \textit{unimodal} framework, as displayed in Fig.~\ref{fig3}(a), focusing solely on \textit{image information} \cite{hendrycks2020pretrained,zhang2024vision}. While effective to a degree, these approaches were inherently limited due to the absence of supplementary contextual information from other modalities \cite{ming2022delving,jiangnegative2024neg}. In recent developments, the advent of Contrastive Language–Image Pre-trained (CLIP) \cite{radford2021learning}, a representative Vision-Language Model (VLM) trained on large-scale \textit{image-text pairs}, has showcased exceptional potential in addressing computer vision (CV) tasks. In the context of OOD detection, the CLIP-like paradigm has moved beyond the limitations of unimodal paradigm by integrating the additional information from the text modality, as depicted in Fig.~\ref{fig3}(b). This shift has led to a surge of research in recent years, as illustrated in Fig.~\ref{fig2}. One prevailing categorization scheme for existing CLIP-like studies is based on \textit{the availability of ID image} during the training or inference. It divides thses studies into \textit{few-shot} and \textit{zero-shot} categories \cite{wang2023clipn,miyai2024generalized,jiangnegative2024neg,miyai2024locoop,sun2024clip,zhang2024lapt}, where few-shot involves limited access to ID image information, while zero-shot operates without any. However, under this classification scheme, the term \textit{‘unknown’} typically implies \textit{unseen} visual information of data, even though the textual information is actually \textit{known} in certain CLIP-like methods. Thus, it overlooks an essential: CLIP is inherently a multimodal pre-trained model, designed to combine the strengths of both image and text modalities, representing a significant shift from the traditional OOD detection paradigm.

In this survey, we aim to bridge this gap by better aligning the multimodal nature of the CLIP framework and explicitly summarizing the associated methodologies. The rest of the paper is organized as follows. We begin by reviewing the background of OOD detection from fundamental aspects (Sec.~2). Then, we present the problem formulation for CLIP-like OOD detection methods (Sec.~3). In contrast to the \textit{few-shot} and \textit{zero-shot} types, we categorize the existing CLIP-like OOD detection works based on \textit{how OOD data are utilized within multimodal settings} under two training strategies (\textit{e.g,} train-free and train-required), and divide them into four groups: OOD Images Seen $+$ OOD Texts Known, OOD Images Unseen $+$ OOD Texts Known, OOD Images seen $+$ OOD Texts Unknown, and OOD Images Unseen $+$ OOD Texts Unknown (\textit{i.e.,} notably, \textit{images seen} represents the outliers rather than real OOD data) (Sec.~4). Furthermore, we also discuss several open questions and limitations of CLIP-like OOD detection and highlight promising directions for future research at cross-domain integration, practical applications, and theoretical understanding aspects (Sec.~5). 

\section{Reviewing the OOD Detection}

The concept of OOD detection has evolved from related topics such as anomaly detection (AD), open-set recognition (OSR) and novelty detection (ND), with its origins tracing back to early studies in statistics and machine learning. To the best of our knowledge, \cite{hendrycks2016baseline} was the first to formally introduce the term \textit{Out-of-Distribution Detection} and explore it in the deep learning era, establishing the foundation for subsequent research. The following subsections will provide an overview of OOD detection from various perspectives.

\subsection{Definition of OOD Detection. }



\begin{definition}[\textbf{Unimodal Out-of-distribution Detection}] 
    \label{def ood}
    Let $\mathcal{X}_{in}$ and $\mathcal{Y}_{in}$ denote ID image space and ID label space, $\mathcal{X}_{ood}$ and $\mathcal{Y}_{ood}$ denote OOD image space and OOD label space, where $ \mathcal{Y}_{in} \cap \mathcal{Y}_{ood} = \emptyset$. Given any text sample $x$, the OOD detection is introduced to classify $x$ into a correct ID class (i.e., $x \in \mathcal{X}_{in}$) or reject it as the OOD ($x \in \mathcal{X}_{ood}$) with a score function $S(\cdot)$ that satisfies:
    \begin{align}
    G_{\gamma}(x; \mathcal{Y}_{in}, \boldsymbol{I}) = \begin{cases} 
    \text{ID}, &\text{if } S(x; \mathcal{Y}_{in}, \boldsymbol{I}) \ge \gamma,\\ 
    \text{OOD},  &\text{othersise}.
    \label{eq1}
    \end{cases}
    \end{align}

    where $G_{\gamma}$ is the OOD detector with the threshold $\gamma$, and $\boldsymbol{I}$ is an image encoder.
    
    
\end{definition}

\subsection{Categorization of Unimodal OOD Detection. } 
The design of OOD detection algorithms involves multiple aspects, such as training strategy, score design, and data types. Since we will categorize the CLIP-like methods based on the utilization differences of OOD data, we briefly review unimodal approaches in this subsection through outliers-exposure and outliers-free \cite{zhang2024vision}: 

\quad \textit{Outliers-exposure} allows partial exposure of OOD data during the training process to attain a more discriminative classification boundary between ID and OOD classes. A key principle is that the exposed OOD data used in training must not share the same semantic distribution with the OOD data used in testing (\textit{i.e.,} $\mathcal{Y}_{outliers} \cap \mathcal{Y}_{ood_{test}} = \emptyset$). Specifically, these outliers may include subsets from real OOD data (\textit{i.e., }such as the background \cite{cho2022towards}), OOD samples from additional datasets \cite{ming2022poem} and generated data \cite{li2024all}. In general, outliers-exposure methods exhibit superior performance because more data information is accessible during training. However, they are sensitive to the choice of outliers, and some generative approaches may introduce significant model complexity and run-time consumption \cite{yang2024generalized}.

\quad \textit{Outliers-free} methods can be broadly categorized into classification-based, density-based and distance-based approaches \cite{yang2024generalized}. Among classification-based methods, post-hoc detection is a major branch that can be directly applied to any already trained models \cite{hendrycks2016baseline,liu2020energy}. Besides, confidence enhancement is another promising branch of classification-based methods, which typically requires model retraining to enhance the robustness of models \cite{bitterwolf2020certifiably}. Furthermore, density-based methods rely on probabilistic models to represent the in-distribution and classify test samples as OOD if they fall into low-density regions \cite{ren2019likelihood,xiao2020likelihood}. In contrast, distance-based methods \cite{sun2022out} identify OOD samples by calculating their distance from the nearest ID centroids or prototypes in the feature space.


\subsection{Benchmark Datasets \& Evaluation Metrics. }

\noindent \textbf{Benchmark Datasets.} Benchmarking OOD detection generally involves defining one dataset as the ID set and selecting several datasets with no category overlap to serve as the OOD datasets. Benchmarking OOD detection generally involves defining one dataset as the ID set and selecting several datasets with no category overlap with the ID set to serve as the OOD datasets. A widely used benchmark for OOD detection is the ImageNet benchmark. When using the large-scale ImageNet-1K dataset as the ID set, iNaturalist, SUN, Texture, and Places are commonly employed as OOD datasets to evaluate algorithm performance. These datasets undergo rigorous filtering to ensure that no samples overlap with the ID classes. Recently, some studies have further divided OOD datasets into near-OOD (e.g., iNaturalist, Species, OpenImage-O, ImageNet-O, SSB-hard and NINCO, \textit{etc}) and far-OOD (Texture, MNIST and SVHN, \textit{etc}) based on the semantic similarity to the ID dataset (\textit{e.g.,} ImageNet-1K). The sources of these datasets are documented in \cite{yang2022openood} and \cite{zhang2024lapt}.

\noindent \textbf{Evaluation Metrics.} We present four evaluation metrics widely used in OOD detection. \textbf{FPR@95} quantifies the false positive rate (FPR) of OOD samples when the true positive rate (TPR) of ID samples is fixed at 95\%. A lower FPR@95 value signifies better OOD detection performance. \textbf{AUROC} the Area Under the Receiver Operating Characteristic Curve (AUROC) represents the relationship between the TPR and the FPR. A higher AUROC value reflects superior model performance. \textbf{AUPR} is a metric that measures the area under the Precision-Recall (PR) curve, which illustrates the relationship between precision and recall. \textbf{ACC} reflects the accuracy of classification for ID classes, with higher values indicating better performance.

\section{Problem Formulation for CLIP-like OOD Detection}

\subsection{CLIP-like Models. } 
CLIP is one of the most advanced pre-trained vision-language models (VLMs) developed by \cite{radford2021learning} from OpenAI. It is designed to align image-text pairs from large-scale datasets within a shared vector space using contrastive learning. This approach allows the model to understand and process the associations between images and text within a latent space.

CLIP is composed of an image encoder $\boldsymbol{I}$ and a text encoder $\boldsymbol{T}$. For a given test image $\boldsymbol{x}$ and its associated label $y_k \in \mathcal{Y}$, CLIP-like modes extract the image embedding $\boldsymbol{v}\in\mathbb{R}^D$ and text embedding $\boldsymbol{t}\in\mathbb{R}^D$ as follows:
\vspace{0.15em}
\begin{align}
    \boldsymbol{v}= \boldsymbol{I(x)},\quad \boldsymbol{t}=\boldsymbol{T}(\boldsymbol{\mathrm{prompt}}(y_k)), k=1,2, ...,K,
\end{align} 
where $D$ denotes the embedding dimension, and $\boldsymbol{\mathrm{prompt}}(\cdot)$ refers to the text prompt function, which is typically crafted manually as ‘a photo of a $[LABEL]$’. For instance, the $[LABEL]$ token can be substituted with specific class names like \textit{'dog'} or \textit{'cat'}. The CLIP-like models then performs zero-shot classification by calculating the cosine similarity between the image embedding $\boldsymbol{v}$ and text embeddings $\boldsymbol{t}_{1}, \boldsymbol{t}_{2}, ..., \boldsymbol{t}_{K}$, scaled by a factor $\tau$:
\begin{equation}
    p\left(y_k \mid \boldsymbol{x} \right)= \frac{\exp\left(\cos \left(\boldsymbol{v},  \boldsymbol{t}_k \right)/ \tau \right)}{\sum_{j=1}^K \exp \left( \cos\left(\boldsymbol{v}, \boldsymbol{t}_j\right)/ \tau \right)}.
\end{equation}

\subsection{CLIP-based Prompt Learning. }
Recent efforts have investigated diverse strategies to further enhance the CLIP’s performance on downstream tasks \cite{li2024learning}, with a major focus on optimizing the text modality. While basic CLIP models use manually crafted prompt templates like ‘a photo of a [LABEL]’, CoOp \cite{zhou2022learning} introduces prompt learning, where the continuous learnable tensors are designed towards the embedding layer of the prompts. Specifically, these tensors are initialized as $\boldsymbol{\omega}_k=[\mathrm{\omega}]_1[\mathrm{\omega}]_2\ldots[\mathrm{\omega}]_L[\mathrm{LABEL}_k]$, where $L$ denotes the token length and $[\mathrm{LABEL}_k]$ is the word embedding of the $k$-th class name. Then, the encoders of CLIP are frozen, and optimization is performed through backpropagation by minimizing classification loss on the target task with few data.

Given the prompt $\boldsymbol{\omega}_k$ as input, the text encoder $\boldsymbol{T}$ outputs the textual embedding as $\boldsymbol{t}_k = \boldsymbol{T}(\boldsymbol{\omega}_k)$, and the final prediction probability is then calculated based on Eq.~(3).

\subsection{CLIP-like OOD Detection. }
The pipeline of CLIP-like OOD detection can been seen in Fig.~\ref{fig3}(b). In particular, the text embedding $\boldsymbol{t}$ extracted by the text encoder $\boldsymbol{T}$ in CLIP can be regarded as a cosine similarity-based classifier, outputting logits or softmax probabilities. One of the various score functions $S(\cdot)$ is then designed towards these outputs, with a threshold $\gamma$ applied to determine whether a test sample $x$ belongs to the ID or OOD category:
    \begin{align}
    G_{\gamma}(x; \mathcal{Y}_{in}, \boldsymbol{I}, \boldsymbol{T}) = \begin{cases} 
    \text{ID}, &\text{if } S(x; \mathcal{Y}_{in}, \boldsymbol{I}, \boldsymbol{T}) \ge \gamma,\\ 
    \text{OOD},  &\text{othersise}.
    \label{eq4}
    \end{cases}
    \end{align}

For samples identified as ID, the class prediction can be determined by selecting the nearest text embedding $\boldsymbol{t}_k$. What sets Eq.~(\ref{eq4}) apart from formulation Eq.~(\ref{eq1}) most distinctly is the integration of encoder $\boldsymbol{T}$, which plays a crucial role in its overall framework.


\begin{figure*}[t]
\centering
\includegraphics[scale=0.66]{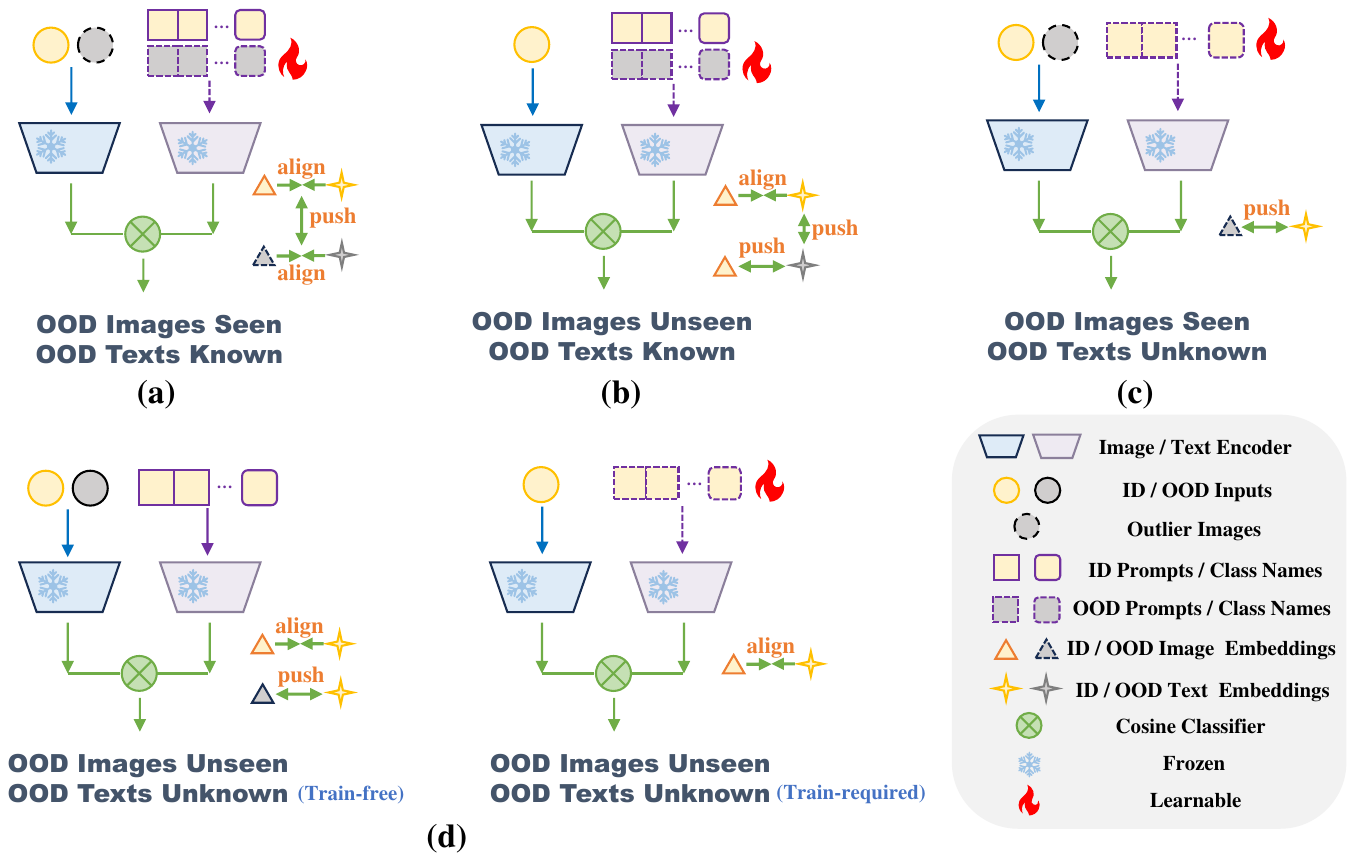} 
\caption{An illustration of different CLIP-like OOD detection paradigms, from the perspective of the utilization of OOD visual and textual information. These paradigms are mainly distinguished by whether they require additional OOD image information (represented by gray dashed circles) and whether OOD-specific textual information (represented by gray dashed squares) is incorporated into fine-tuning. In particular, \textit{OOD Images Unseen + OOD Texts Unknown} is further divided into two groups, \textit{Train-free} and \textit{Train-required} (illustrated in Figure (d) with \textcolor{blue}{blue descriptions}). The primary criterion for this division is whether prompt learning for ID classes is involved.}
\label{fig4}
\end{figure*}

\section{CLIP-like OOD Detection: Categorization}
In this section, we introduce the methodologies for CLIP-like OOD detection. We will categorize recent studies based on \textit{how visual and textual information of OOD data are utilized in multimodal settings} under train-free and train-required strategies. Specifically, building on CLIP's multimodal nature, we analyze various design strategies for \textit{OOD data} across image and text modalities and systematically classify existing methods into four categories: \textit{Images Seen $+$ Texts Known, Images Unseen $+$ Texts Known, Images Seen $+$ Texts Unknown, and Images Unseen $+$ Texts Unknown}. These categories are summarized in Table.~\ref{tab:1}, with their corresponding frameworks illustrated in Fig.~\ref{fig4}.

Notably, in this paper, the term \textit{OOD Images Seen} often refers to \textit{train-required} strategy. These OOD images do not correspond to the actual OOD samples used in testing.

\subsection{OOD Images Seen \texorpdfstring{$+$}{+} Texts Known}

Studies in this category typically involve incorporating outliers and constructing specific prompts for fine-tuning (\textit{a.k.a} prompt learning). ID-like \cite{bai2024id} generates outliers by randomly cropping a small subset of ID data and introduces a prompt learning framework tailored for these ID-like outliers. CLIP-driven \cite{sun2024clip} generates reliable OOD data by mixing ID-related features with an unknown-aware prompt learning strategy. CATEX \cite{liu2024category} synthesizes outliers independent of specific ID tasks by applying random perturbations, aiding in training non-trivial spurious text descriptions. LAPT \cite{zhang2024lapt} also employs the mix-up technique to synthesize outliers, exploring the space between the ID and OOD distributions. Furthermore, it expands the OOD label space by NegLabel \cite{jiangnegative2024neg} and enables label-driven automated prompt tuning via web-scale retrieval and text-to-image generation techniques. To resolve semantic misalignment caused by uniform negative labels across different OOD datasets, AdaNeg \cite{zhang2024adaneg} adaptively generates outliers by exploring actual OOD images stored in a purpose-built memory bank, achieving better alignment with the OOD label space.

\textbf{Remarks:} This category is defined by two main aspects: 1) \textbf{Images Seen:} ID images are used to generate outlier images through techniques like image transformations or mix-up. These outlier images are then utilized for model fine-tuning (\textit{i.e.,} train-required); 2) \textbf{Texts Known:} OOD text (\textit{i.e., prompt}) is initialized initially and combined with the generated outlier images in prompt learning, enhancing the model’s ability to distinguish between ID and OOD samples. Notably, AdaNeg is the only method that does not involve prompt learning, however, its designed memory bank is still involved in the training process.

\subsection{OOD Images Unseen \texorpdfstring{$+$}{+} Texts Known}

This category constitutes a major branch of CLIP-based OOD detection methods, fully exploiting CLIP's textual interpretability. Zero-shot OE \cite{fort2021exploring}, which selects the candidate labels related to OOD classes from an ID labels-irrelevant dataset, then applies Softmax to compute prediction probabilities. ZOC \cite{esmaeilpour2022zero} expands the OOD label space by employing a text generator on top of CLIP to select candidate OOD class names for each test sample. CLIPN \cite{wang2023clipn} introduces a learnable “no” prompt and a specialized text encoder to capture negative semantics within ID images. Textual OE \cite{park2023powerfulness} emphasizes the text modality, highlighting the benefits of textual outliers by generating OOD class names through three strategies: word-level, description-level and caption-level textual outliers. NegLabel \cite{jiangnegative2024neg} derives a large set of negative labels from extensive corpus databases, utilizing the distance between negative labels and ID labels as a metric. Additionally, it designs a novel OOD scoring scheme collaborated with negative labels. EOE \cite{caoen2024visioning} constructs OOD texts to generate potential outlier class labels for OOD detection. For instance, it employs prompts like \textit{“Give three categories that visually resemble a horse”}to align with the ID class \textit{“horse”}. It also introduces a new score function based on these potential outliers. NegPrompt \cite{li2024learning} trains a set of negative prompts exclusively on ID data, where each prompt represents a negative interpretation of a specific ID class label, helping define the boundary between ID and OOD images. Unlike conventional prompt learning, which learns positive prompts for all classes, LSN \cite{nie2024out} focuses on learning negative prompts for each OOD class to explicitly define what is \textit{“not”}. CMA \cite{lee2024cma} employs neutral prompts as the agents to augment the label space, which does not apply learnable prompts for text enhancement, but only uses the corresponding class names.

\textbf{Remarks: }This category has two main characteristics: 1) \textbf{Images Unseen: }No additional OOD images are needed for training or testing; 2) \textbf{Texts Known: }The primary focus of this category is on OOD text design, which can be further divided into: a) \textit{\textbf{Learnable OOD prompts design.}} Methods in this group, such as CLIPN, LSN, NegPrompt and CMA, construct a variety of learnable prompts with \textit{“no/not”} semantic logic, typically derived from the textual information of ID data; b) \textit{\textbf{OOD class names mining.}} This branch, including Zero-shot OE, ZOC, Textual OE, NegLabel and EOE, focuses on mining OOD class name candidates from external text databases using various predefined metrics. Overall, most methods in this category require training to optimize the designed OOD textual information (\textit{i.e., }train-required), except for Zero-shot OE, NegLabel and CMA(\textit{i.e., }train-free).



\subsection{OOD Images Seen \texorpdfstring{$+$}{+} Texts Unknown}

Unlike the \textit{OOD Images Seen $+$ OOD Texts Known}, this category does not rely on enhancing the similarity between designed OOD texts and OOD images. Instead, it aims to maximize the dissimilarity between ID texts and OOD images through various OOD regularization techniques. LoCoOp \cite{miyai2024locoop} first segments original ID images into local regions, treating ID-irrelevant areas (\textit{i.e.,} background) as outliers. It then applies entropy maximization as an OOD regularization strategy to ensure that features in ID-irrelevant regions remain separate from any ID text embeddings. Another work, TagFog \cite{chen2024tagfog} introduces the jigsaw-based fake OOD images as outliers and incorporates rich semantic embeddings from ChatGPT-generated description of ID textual information to guide the training of the image encoder.

\textbf{Remarks: }This category has two defining characteristics: 1) \textit{Images Seen: }Similar to \textit{OOD Images Seen $+$ OOD Texts Known}, it involves acquiring external images as outliers through various techniques for training (\textit{i.e., }train-required); 2) \textbf{Texts Unknown: } As previously mentioned, these methods apply OOD regularization to separate the obtained outliers from ID class text embeddings, thereby effectively removing redundant information from ID text representations.

\subsection{OOD Images Unseen \texorpdfstring{$+$}{+} Texts Unknown}



One subcategory adopts a train-free strategy, also referred to as test-time OOD detection in CLIP-like models, corresponding to post-hoc methods in unimodal OOD detection. It does not require training but instead employs designed OOD score functions at test time to recognize OOD data. A key representative work is MCM \cite{ming2022delving}, which introduces the Maximum Concept Matching score. This method considers ID class names as textual concepts that align with visual embeddings during testing, supported by detailed analysis and theoretical insights. GL-MCM \cite{miyai2023zero} improves on MCM by incorporating local features, though its applicability is restricted to specific scenarios. DPM-F \cite{zhang2024vision} argues that ID visual information plays a crucial role in model decision-making during testing. It stores ID-specific text features as the textual patterns and integrates them with ID visual information as visual patterns to aid in OOD detection. SeTAR \cite{li2024setar} extends MCM by introducing selective low-rank approximations to improve performance. TAG \cite{liu2024tag} augments the label space by using only ID label information to improve OOD detection.

\begin{table}[htbp]
  \centering
    \begin{tabular}{c|c|p{11em}}
    \toprule
    Categories & Train Types & \multicolumn{1}{c}{Methods} \\    
    \midrule
     \makecell{IS+TK} & \makecell{TR}  & \makecell{ID-like, LAPT, AdaNeg, \\ CLIP-driven, CATEX} \\
    \midrule
     \multirow{4}[0]{*}{IU+TK} &  \makecell{TR} &  \makecell{ZOC, EOE, CLIPN, LSN, \\NegPrompt, Textual OE}  \\
    \cmidrule{2-3}  &  \makecell{TF} & \makecell{ZerO-shot OE, CMA, \\ NegLabel} \\
    \midrule
    IS+TU & TR    & \multicolumn{1}{c}{LoCoOp, TagFog} \\
    \midrule
    \multirow{3}[3]{*}{IU+TU} & TR    & \multicolumn{1}{c}{GalLop, DPM-T} \\
    \cmidrule{2-3}  &  \makecell{TF}    &  \makecell{MCM, GL-MCM, SeTAR, \\ DPM-F, TAG} \\ 
    \bottomrule
    \end{tabular}
    \caption{Different Categories for CLIP-like OOD Detection. Where TR and TF refers to train-required and train-free, respectively. The abbreviations in the Categories are consistent with those in Fig.~\ref{fig2}.}
  \label{tab:1}
\end{table}%

Another subcategory adopts the train-required strategy, eliminating the need to design OOD scores during testing. GalLoP \cite{lafon2024gallop} also does not require information from OOD images or OOD texts. Instead of designing OOD scores, it focuses on ID images at both the global and local levels, separately designing and training ID prompts (\textit{i.e., }train-required) to achieve fine-grained and precise text-to-image matching. DPM-T \cite{zhang2024vision} improves performance by further fine-tuning the ID prompts within the original framework.

\textbf{Remarks: }This category has two key characteristics: 1) \textbf{Images Unseen: }None of the methods in this category require additional OOD images during training or inference, regardless of whether they are train-free or train-required strategies. Furthermore, the designed OOD scores can be applied across various CLIP-like OOD detection approaches; 2) \textbf{Texts Unknown: } In train-free methods, prompt learning is unnecessary because OOD scores are only used during inference. In train-required methods, only ID texts are learned, and there is no requirement for OOD texts.

\subsection{Advantages Comparisons}

The pipelines for the four categories are illustrated in Fig.~\ref{fig4}. \textit{OOD Images Seen \texorpdfstring{$+$}{+} Texts Known} integrates additional information from both vision and text, allowing the model to more effectively identify OOD data. By incorporating prior knowledge of OOD images and textual descriptions, the model can learn richer representations that enhance detection. \textit{OOD Images Unseen \texorpdfstring{$+$}{+} Texts Known} takes a different approach by avoiding the explicit introduction of additional OOD samples. Instead, it focuses on designing learnable OOD-specific vectors or class labels that encourage the model to push the negative OOD texts away from ID visual features. \textit{OOD Images Seen \texorpdfstring{$+$}{+} Texts Unknown} offers an opportunity to explore various generative strategies for creating effective OOD samples, making it a valuable testbed for investigating the impact of synthetic data in OOD detection. \textit{OOD Images Unseen \texorpdfstring{$+$}{+} Texts Unknown} adopts a minimalistic approach by forgoing both outlier generation and OOD-specific prompt learning. This ensures a computationally efficient pipeline, making it particularly suitable for scenarios with limited computational resources.

\section{Limitations and Future Directions}

\subsection{Limitations}

\noindent \textbf{OOD Information Leakage. }
While this concern is widely acknowledged, it remains crucial for the advancement of CLIP-like OOD detection. CLIP is trained using contrastive learning on large-scale image-text datasets, such as LAION-400M. As a result, during pre-training, CLIP becomes exposed to and learns certain features of samples or categories that may appear in the test data. The leakage of OOD information can introduce two key challenges: 1) The model may rely on previously learned categories or features, misclassifying real OOD samples as part of an ID category; 2) Some \textit{OOD Texts Known} methods that incorporate “no” or “not” semantics, may explicitly identify OOD data during testing instead of genuinely inferring them. This can result in artificially inflated performance, which may lead to issues when applied in real-world scenarios. Therefore, further exploration of solutions, such as refining pre-trained data, introducing specialized mechanisms, and designing tailored training strategies, is necessary to mitigate this issue.

\noindent \textbf{Evaluation Benchmarks. }
This issue stems from information leakage. Current CLIP-like OOD detection methods often rely on commonly seen object categories during testing, leading to inevitable semantic overlap with CLIP’s pre-trained classes. Existing benchmark datasets require further improvement in task difficulty and category diversity. To truly assess the real detection capability of CLIP-like OOD detection, test datasets should be replaced with those from more specialized domains, such as Pathology Image Datasets (\textit{e.g.,} CAMELYON16 \& CAMELYON17, DigestPath 2019) or Geospatial Remote Sensing Datasets (\textit{e.g.,} EuroSAT and BigEarthNet).

\noindent \textbf{Data Noise Issue. }
Given the diverse and inconsistent quality of data sources, CLIP may encounter image and label noise when pre-training on large web-scale datasets, making it vulnerable to data noise. For instance, image noise arises from low-quality images, atypical perspectives, blurriness, or occluded scenes. Label noise occurs when certain image-text pairs are imperfectly matched, such as a direct semantic mismatch where an image of a cat is incorrectly labeled as a dog. Such data noise issue could distort the features learned by CLIP, ultimately degrading its ability to distinguish OOD samples during testing.

\subsection{Future Directions}

\noindent \textbf{Combining with More Complex and Dynamic Environments. }We believe that CLIP's exceptional capabilities extend beyond the well-explored domain of OOD detection. Moreover, there remains substantial untapped potential in combining OOD detection with more complex and dynamic scenarios. For example: 1) \textit{CLIP-like OOD in Long-Tailed Recognition.} In extremely imbalanced long-tailed distributions, the models may struggle to capture sufficient features of minority classes, erroneously treating these samples as OOD data. We propose addressing this issue by utilizing CLIP's powerful textual description capabilities, which allow for the creation of more specific and detailed descriptions for tail classes; 2) \textit{OOD Challenges in Continual Learning with CLIP.} Catastrophic forgetting is a fundamental challenge in continual learning. As the model adapts to new tasks, it may forget the knowledge from previous tasks. However, traditional OOD detection methods often rely on features extracted from past tasks to establish decision boundaries. Therefore, we think that, by using CLIP-based similarity sampling, we can select representative historical data for replay, and store representative samples from previous tasks to prevent the model from forgetting the previous OOD boundaries; 3) \textit{CLIP-like Multi-label OOD Detection.} Current CLIP-like OOD detection methods are largely confined to single-label tasks. However, real-world scenarios often involve multi-label images. In multi-label tasks, a single sample may simultaneously contain both ID and OOD labels. Effectively distinguishing between ID and OOD labels in multi-label tasks is a crucial research direction.


\noindent \textbf{Diverse Real Scenarios. }
As discussed in Sec.~5.1, CLIP exhibits robust classification and OOD detection capabilities for natural images. However, its potential in specialized domains remains largely unexplored. Here, we highlight some real-world applications: 1) \textit{Intelligent Mine Safety: }Underground mining environments are confined spaces with intricate conditions. Strict safety regulations often restrict the deployment of electronic devices such as sensors. However, inadequate monitoring equipment may fail to detect early warning signs of mining accidents. Consequently, harnessing CLIP’s robust generalization capabilities, we seek to identify environmental anomalies (OOD) that deviate from normal safety conditions (ID), enabling proactive risk mitigation and emergency preparedness; 2) \textit{Medical Electrocardiogram (ECG): }Current ECG analysis systems mainly diagnose common conditions such as arrhythmias and heart disease. However, certain anomalies in ECG signals may not be included in standard training datasets, rendering unimodal detection methods ineffective. With the contrastive learning mechanism of CLIP, it becomes possible to identify even minor variations in ECG signals. By comparing these signals with abnormal descriptions provided by medical professionals, the system can assess whether an anomaly is OOD and potentially detect rare or previously unknown diseases.

\noindent \textbf{More Advanced Theories. }Most current research on CLIP-like OOD detection primarily focuses on practical applications, while its theoretical understanding remains relatively underexplored. One theoretical perspective that merits deeper investigation is the concept of the \textit{Modality Gap} \cite{liang2022mind}. This phenomenon describes an observation that after being processed by CLIP’s respective encoders, image embeddings and text embeddings tend to occupy distinct and non-overlapping regions within the shared feature space. While some CLIP-like OOD detection methods advocate for bridging this gap to improve performance \cite{zhang2024vision}, the necessity and desirability of eliminating the modality gap remain uncertain. As a result, despite CLIP’s empirical success in the OOD detection task, a more rigorous theoretical analysis is required to understand its underlying mechanisms and limitations.



\section*{Acknowledgments}
This work was supported by the National Natural Science Foundation of China (Grant No. 62376126, 62106102), the Hong Kong Scholars Program (Grant No. XJ2023035) and the Fundamental Research Funds for the Central Universities (Grant No. NS2024058).

\clearpage
\bibliographystyle{named}
\bibliography{ijcai25}

\end{document}